\documentclass[preprint,12pt]{elsarticle}
\usepackage{epsfig}
\usepackage{amssymb}
\usepackage{amsmath}
\usepackage{amsthm}
\usepackage{bm}
\usepackage{caption}
\usepackage{subcaption}
\usepackage{mathabx}
\newcommand{\ignore}[1]{}
\usepackage{lipsum}
\makeatletter
\def\ps@pprintTitle{%
 \let\@oddhead\@empty
 \let\@evenhead\@empty
 \def\@oddfoot{}%
 \let \@oddfoot}
\makeatother
\begin{document}

\begin{frontmatter}

\title{An Interpretable Generative Model for Handwritten Digit Image Synthesis}

\author{Yao Zhu\textsuperscript{1}, Saksham Suri\textsuperscript{2},
	Pranav Kulkarni\textsuperscript{3}, Yueru Chen\textsuperscript{1}, Jiali Duan\textsuperscript{1} and C.-C. Jay Kuo\textsuperscript{1}}
\address{\textsuperscript{1}University of Southern California, Los
	Angeles, CA, USA, \textsuperscript{2}IIIT, Delhi, India,
	\textsuperscript{3}IIT, Mumbai, India}

\begin{abstract}

An interpretable generative model for handwritten digits synthesis is
proposed in this work. Modern image generative models, such as
Generative Adversarial Networks (GANs) and Variational Autoencoders
(VAEs), are trained by backpropagation (BP). The training process is
complex and the underlying mechanism is difficult to explain. We propose
an interpretable multi-stage PCA method to achieve the same goal and use
handwritten digit images synthesis as an illustrative example. First, we
derive principal-component-analysis-based (PCA-based) transform kernels
at each stage based on the covariance of its inputs. This results in a
sequence of transforms that convert input images of correlated pixels to
spectral vectors of uncorrelated components. In other words, it is a
whitening process. Then, we can synthesize an image based on random
vectors and multi-stage transform kernels through a coloring process.
The generative model is a feedforward (FF) design since no BP is used in
model parameter determination. Its design complexity is significantly
lower, and the whole design process is explainable. Finally, we design
an FF generative model using the MNIST dataset, compare synthesis
results with those obtained by state-of-the-art GAN and VAE methods, and
show that the proposed generative model achieves comparable performance. 

\begin{keyword}
Generative Model, Feedforward Model, Interpretable Model, Principle Component 
Analysis (PCA), GAN, VAE
\end{keyword}
\end{abstract}

\end{frontmatter}

\section{Introduction}\label{sec:introduction}

Data-driven image synthesis is a popular topic for a variety of
vision-based applications nowadays. A great majority of modern image
synthesis tasks are accomplished by Generative Adversarial Networks
(GANs) and Variational Autoencoders (VAEs). The GAN approach trains two
networks at the same time, namely, a generator and a discriminator. They
are trained with respect to each other adverserially. The generator
takes a random vector as the input and generates an image by adapting
network parameters to make synthesized and real ones as close as
possible. Their difference is detected by the discriminator.  The whole
training process is cast as a multi-layer non-convex optimization
problem and solved by backpropagation (BP).  Although the BP approach
provides reasonable results, the whole process is complex and
mathematically intractable. 

To offer an interpretable image generative model, one idea is to adopt a
one-stage whitening and coloring process using the principal component
analysis (PCA), which is also known as the Karhunen-Lo$\acute{e}$ve
transform (KLT). However, the one-stage transform cannot generate images
of high resolution. To overcome this difficulty, we derive a multi-stage
PCA by cascading multiple transform stages. It relates input images of
correlated pixels to output spectral vectors of uncorrelated components
through a multi-stage whitening process. To accomplish the synthesis
task, we conduct multi-stage transforms in the reversed order that
corresponds to a multi-stage coloring process. The resulting generative
model is called a feedforward one since there is no backpropagation used
in model parameters determination. The complexity of the feedforward
model is low and the whole design process is explainable. 

The rest of the paper is organized as follows. Related previous work is
reviewed in Sec. \ref{sec:related}. One-stage and multi-stage generative
models are presented in Secs. \ref{sec:onestage} and \ref{sec:Main},
respectively. Experimental results are shown in Sec. \ref{sec:enhance}.
Some discussion is made in Sec. \ref{sec:comparative}.  Finally,
concluding remarks are given in Sec. \ref{sec:conclusion}. 

\section{Review of related work}\label{sec:related}

Generative models are an important topic in computer vision and machine
learning. Among many state-of-the-art generative models for image
synthesis \cite{dosovitskiy2015learning, reed2015deep, yang2015weakly},
there are two main stochastic models. They are VAEs and GANs. The VAE
\cite{kingma2013auto, rezende2014stochastic, oord2016pixel,
	zhao2017learning} is a probablistic graphical model that optimizes the
variational lower bound on data likelihood. The GAN system
\cite{goodfellow2014generative} consists of a generator and a
discriminator. The generator is optimized to fool the discriminator, and
the discriminator is optimized to distinguish real and fake images.
There are several GAN variants to improve stability and efficiency
\cite{arjovsky2017towards, arjovsky2017wasserstein,
	gulrajani2017improved, mao2017least, radford2015unsupervised,
	salimans2016improved, zhao2016energy}. Examples include WGAN,
informative GAN, etc. Takuhiro {\em et al.} \cite{kaneko2018generative}
proposed a decision tree controller GAN, which can learn hierarchically
interpretable representations without relying on detailed supervision.
Huang {\em et al.} \cite{huang2018introvae} proposed an introspective
variational autoencoder (IntroVAE) model to synthesize high-resolution
photographic images. It can conduct self-evaluation of generated images
for quality improvement accordingly. 

Research on interpretable generative methods for image synthesis is much
less. One idea is to examine hierarchical representations for image
synthesis.  Examples include: \cite{zhang2017stackgan,
	wang2016generative, vondrick2016generating, huang2017stacked}. Another
idea is to adopt a recursive structure \cite{eslami2016attend,
	gregor2015draw, im2016generating, kwak2016generating, yang2017lr}.
Recently, Kuo {\em et al.} \cite{kuo2016understanding, kuo2017cnn,
	kuo2018data, kuo2018interpretable} studied interpretable convolutional
neural networks (CNNs), and proposed an image synthesis solution based
on the Saak transform in \cite{kuo2018data}. They successfully
reconstructed images using Saak coefficients. Here, we conduct image
synthesis from random vectors rather than image reconstruction, and
present a multi-stage PCA-based generative model for high resolution
image synthesis. 

\section{Generative Model with One-Stage PCA}\label{sec:onestage}

\begin{figure}
	\centering
	\includegraphics[width=10cm]{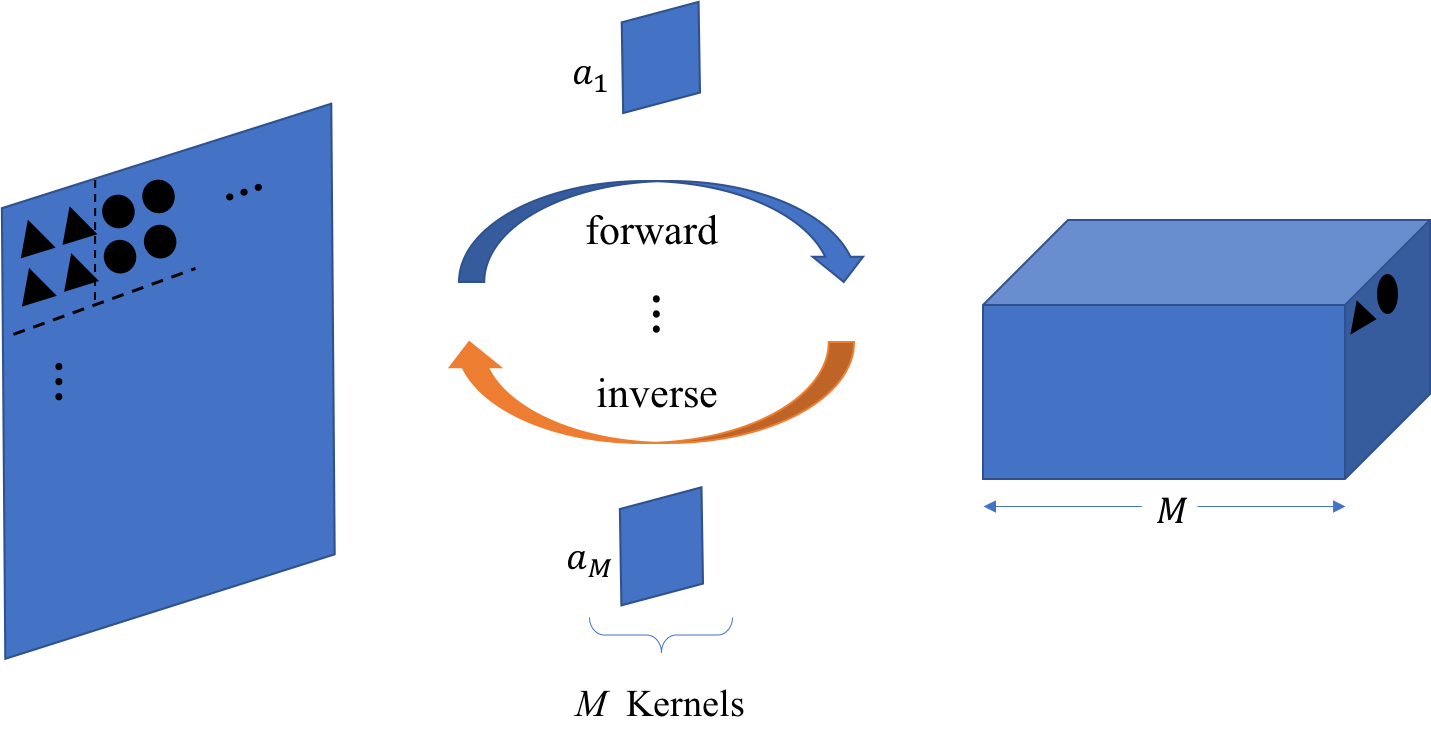}
	\caption{3D representation of single stage PCA transform. Blue flow
		represents forward transform, orange flow represents inverse transform.}
	\label{fig:3Dcuboid}
\end{figure}

The block-diagram of a generative model using the one-stage PCA is shown
in Fig. \ref{fig:3Dcuboid}. It contains a forward transform and an
inverse transform. They correspond to the whitening and the color
processes, respectively. The training data are used to determine the
transform kernels and the distribution of transformed coefficients.
Then, the two pieces of information are used for automatic image
synthesis. 

\noindent
The forward PCA transform consists of the following two steps. 
\begin{enumerate}
	\item Compute transform kernels from input vectors. \\
	By following \cite{kuo2018interpretable}, we decompose input vectors
	into DC (direct current) and AC (alternating current) two components.
	The correlation matrix for AC input vectors is denoted by ${\bf R} \in
	R^{N\times N}$. It has rank $N-1$. That is, its first $N-1$ eigenvalues
	are positive and the last one is zero.  The eigenvectors corresponding
	to positive eigenvalues form an orthonormal basis of the AC subspace.
	Also, we store eigenvectors in a descent order of its corresponding
	eigenvalues. We select the first $M$ (with $M\leq N$) eigenvectors, of
	which its corresponding eigenvalues takes the most energy of the input
	vector space. The first $M$ basis functions, denoted by ${\bf a}_1, {\bf
		a}_2, \cdots , {\bf a}_{M}$ and called transform kernels, provide the
	optimal subspace approximation $R^M$ to $R^N$. 
	
	\item Project AC input ${\bf f}$ to kernels ${\bf a}_1, {\bf a}_2,
	\cdots , {\bf a}_{M}$. \\
	We have projection coefficients shaped in a 3D cuboid, as shown in Fig.
	\ref{fig:3Dcuboid}. The projection coefficients are defined by
	\begin{equation}\label{equa: 1}
	p_i=  {\bf a}_{i}^T {\bf f}, \quad k=1, \cdots, M,
	\end{equation}
	where ${\bf a}_i$, ${\bf f}$ and $M$ denote the $i$th transform kernel,
	the AC input vector and the number of kernels, respectively. 
\end{enumerate}

\noindent
The inverse PCA transform aims to reconstruct the AC input, ${\bf f} \in
R^{N}$, as closely as possible. We use the best linear subspace
approximation to reconstruct the input vector using transform kernels
${\bf b}_{i}$, $i=1, \cdots, M$, and projection coefficient vector 
${\bf q}=(q_1, \cdots, q_M)$. That is, we have
\begin{equation}
\tilde{\bf f} = \sum_{i=1}^{M} q_i {\bf b}_{i}.
\end{equation}
The inverse PCA transform is a mapping from projection vector ${\bf q}$
to an approximation $\tilde{\bf f} \in R^M$ of ${\bf f} \in R^N$. 

It is worthwhile to highlight two points. First, image synthesis is
different from image reconstruction. Image reconstruction attempts to
relate one input image with its transformed coefficients. Image
synthesis intends to relate one class of images with their transform
coefficients. Thus, we should study the distribution of projection
coefficient vector ${\bf p}$. It is well known that the PCA transform is
a whitening process so that elements of ${\bf p}$ are uncorrelated.
Thus, each element can be treated independently. Second, the
single-stage PCA cannot generate high quality images of high resolution
since many PCA components contain high frequency components only.  To
address this problem, we proposed to cascade multiple PCA transforms to
provide a multi-stage generative model. 

\section{Generative Model with Multi-Stage PCA}\label{sec:Main}

\begin{figure*}
	\centering
	{\includegraphics[width=\textwidth]{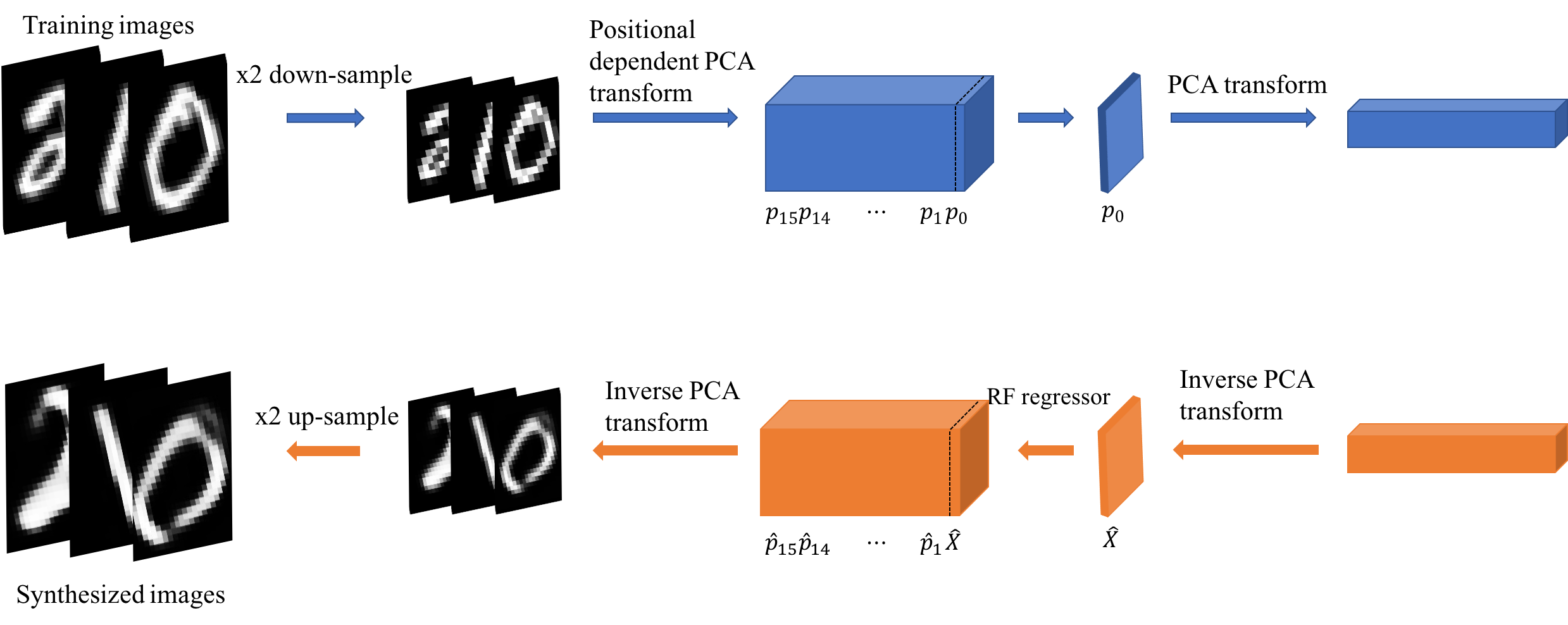}}
	\caption{Illustration of the proposed multi-stage generative model for
		handwritten digits with the MNIST dataset as the training samples, where
		blue arrows represent the training workflow and orange arrows represent
		the synthesis workflow.} \label{fig:overview_1}
\end{figure*}

In this section, we propose a generative model based on multi-stage PCA
transforms.  The block-diagram of the proposed multi-stage generative
model is shown in Fig. \ref{fig:overview_1}. Since the difference
between the original digit images of size $32\times32$ and the
interpolated digit image from size $16 \times 16$ to $32 \times 32$ is
very small, we do an average pooling to reduce all images to size $16
\times 16$ as a pre-processing step to save computational complexity.
Note that this pre-processing step can be removed to allow a more
general processing. We will illustrate this point for more complicated
images such as facial images as future extension. 

\subsection{Multi-Stage Generative Model}\label{ssec:archi}

Our proposed system is a two-stage generative model that consists of two
single-stage PCA transforms, where the spatial resolution at each stage
is $4\times4$. The model parameters are determined in a FF one-pass
manner as follows. 
\begin{itemize}
	\item Stage 1: \\
	Conduct the position-dependent single-stage PCA transform on
	non-overlapping batches of size $4 \times 4 $.  Each position yields PCA
	coefficients of size $16 \times 1 \times 1$, where $16$ is the spectral
	component and $1 \times 1$ is the spatial resolution. In other words,
	the 3D cuboid output has the spatial dimension of each position and a
	spectral dimension of 16.  Among 16 spectral coefficients $p_0, p_1,
	\cdots, p_{15}$, $p_0$ is the DC projection while others are AC
	projections. We only pass the DC projection to the second stage and use
	AC projections of training samples to train a random forest regressor.
	Then, the random forest regressor will be used to predict AC projections
	in the synthesis process. 
	\item Stage 2: \\
	Conduct single-stage PCA transform of dimension $K \times 1 \times1$,
	where $K$ is the spectral dimension and $1 \times1$ is the spatial
	dimension. Thus, the whole input image is transformed to a spectral
	vector of dimension $K$. 
\end{itemize}

The synthesis process is formed by the cascade of two inverse
transforms.  The input vector is a random vector which has the mean and
variance information of PCA coefficients of the last stage.  It
introduces randomness into the synthesis process. 

\subsection{Outlier Detection}\label{sssec: outlier}

In the synthesis procedure, a random vector is used as the start point.
We may see a synthesized vector that lies in the tail region of the
Gaussian distribution. They are outliers and will lead to bad generated
samples in the end. We use two methods to detect outliers and remove
them. One is k-mean clustering combined with the mean squared error. The
other is the Z-score method. 
\begin{itemize}
	\item {\bf K-mean clustering and MSE.} \\
	We apply K-mean clustering on coefficients of the second stage PCA. Each
	cluster has its centroid and the mean squared error. The latter
	indicates the distance between samples and their centroid. We use
	$C_{i,k}$ to denote the coefficient of the $i^{th}$ input image in the
	$k^{th}$ cluster.  The centroid for cluster $k$ is $C_{k}^{*}$.  Then,
	we have the mean squared error of cluster $k$:
	\begin{equation}\label{equa: 3}
	MSE_{k} = \frac{1}{N_k}\sum_{C_{i,k} \in cluster \:k} (C_{i,k} - C_{k}^{*})^2 
	\end{equation}
	For each generated $\hat{C}$, we check which cluster it belongs to. If
	the distance between $\hat{C}$ and its centroid $k$ is greater than
	$MSE_k$, we view it as an outlier, vice versa. 
	\item {\bf Z score.} \\
	The Z score is often used to measure the number of standard deviations
	for a data point to be away from its mean. Here, we use it to detect
	outliers. Since $C_i$ lies in a high dimensional space, we obtain mean
	$\mu$ and standard deviation $\sigma$ of each dimension. We consider
	$C_i$ as an inlier only if all the components of $C_i$ lies within
	$3\sigma$ range of its distribution. 
\end{itemize}
We discard outliers in the image synthesis process.

\subsection{AC Prediction}\label{sssec: condi}

For the model in Fig. \ref{fig:overview_1}, the conditional probability
of each AC projection given the DC projection, i.e. $P(p_i|p_0)$, is
learned in the training procedure. Thus, in the synthesis flow, once we
obtain the generated DC projection, $\hat{X}$, it will be used to
predict AC projections through the inverse transform at the second
stage. The random forest regressor offers a powerful method in
predicting AC projections in two aspects. First, given a set of data
points, both input $X$ and output ${\bf b}_1, {\bf b}_2, \cdots, {\bf
	b}_{M}$ are high dimensional, the random forest regressor works well for
the dataset.  Second, it meets our need. Given $\hat{X}$, the random
forest regressor can give the corresponding ${\bf b}_1, {\bf b}_2,
\cdot, {\bf b}_M$. 

\section{Experimental Results}\label{sec:enhance}


\subsection{Number of Principle Components}\label{ssec:numpca}

For input images of size $\sqrt{N} \times \sqrt{N}$, the total number of
principle components is $N$.  We choose the first $M$ (with $M < N$)
eigenvectors of the correlation matrix to provide the optimal linear
subspace approximation, $R^M$ to the original space, $R^N$. One key
question is the selection of parameter $M$. Here, we use the first-stage
PCA for the MNIST dataset as an example.  

\begin{figure}
	\centering
	{\includegraphics[width=10cm]{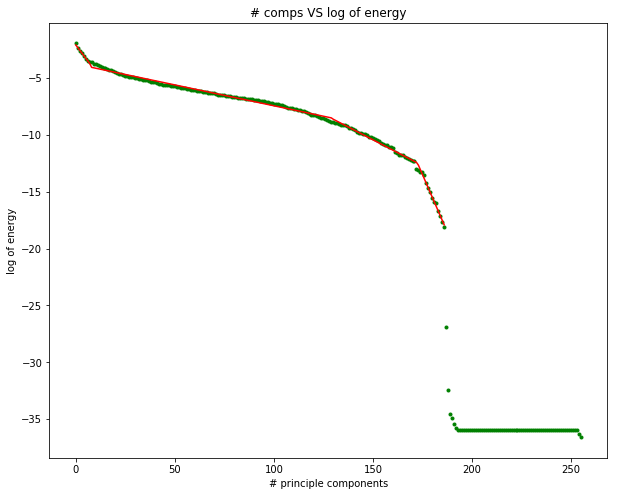}}
	\caption{The energy curve as a function of principal component indices
		for the MNIST dataset.}\label{fig:8energy}
\end{figure}

We show the semi-log energy plot of principle components of input images
in Fig. \ref{fig:8energy}, where the y-axis indicates the ratio of an
eigenvalue $e_i$ and the sum of all eigenvalues $\sum e_i$, $i=1,2,
\cdots, N$, and the x-axis is the indices of principal components.  As
shown in Fig. \ref{fig:8energy}, the whole energy curve can be
decomposed into five linear sectors with four turning points as
separators. The four turning points are given in Table
\ref{Tab:turning}. To derive these turning points automatically, we use
the least square regressor to fit leading data points in one sector and
choose the point that starts to deviate from a straight line segment as
the turning point. 

\begin{table}
	\centering
	\caption{The indices of turning points.}\label{Tab:turning}
	\begin{tabular}{|c||c|c|c|c|} \hline 
		Turning point & 1st & 2nd & 3rd & 4th \\ \hline 
		index & 8 & 120 & 180 & 220 \\ \hline 
	\end{tabular} 
\end{table}

\begin{figure}
	\centering
	\begin{minipage}[b]{.3\linewidth}
		\centering
		{\includegraphics[width=\textwidth]{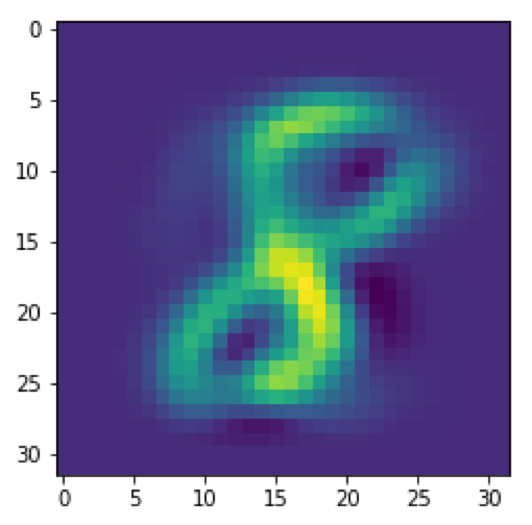}}
		{(a) Section 1}\medskip
		\label{fig:sec_1}
	\end{minipage}
	\hfill
	\begin{minipage}[b]{.3\linewidth}
		\centering
		{\includegraphics[width=\textwidth]{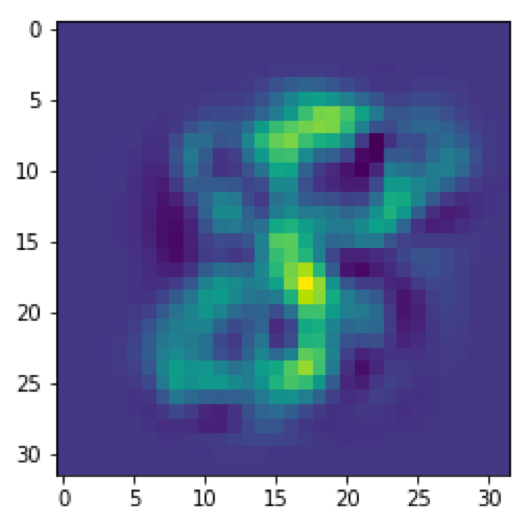}}
		{(b) Section 2}\medskip
		\label{fig:sec_2}
	\end{minipage}
	\hfill
	\begin{minipage}[b]{.3\linewidth}
		\centering
		{\includegraphics[width=\textwidth]{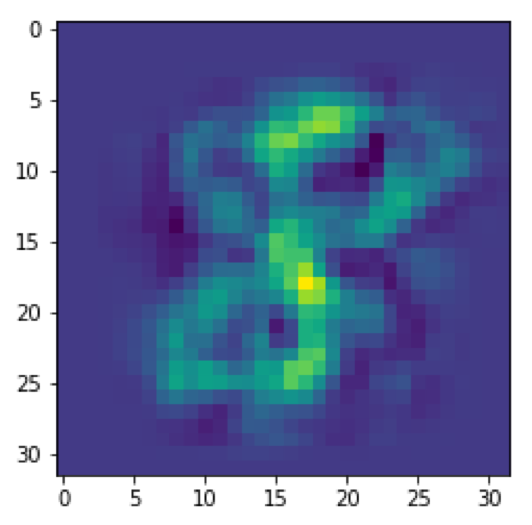}}
		{(c) Section 3}\medskip
		\label{fig:sec_3}
	\end{minipage}
	\caption{Generated images using principle components in different sections.}
	\label{fig:diff_sec}
\end{figure}

\begin{figure}
	\centering
	\begin{minipage}[t]{0.32\linewidth}
		\centering
		\includegraphics[width=\textwidth]{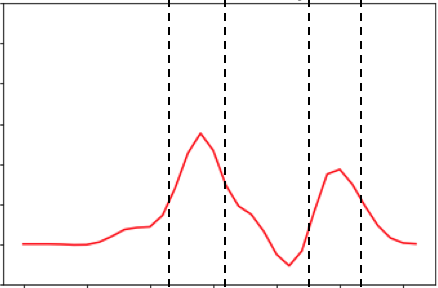}
		(a) section 1\medskip
		\label{fig:hist_1}
	\end{minipage}
	\hfill
	\begin{minipage}[t]{0.32\linewidth}
		\centering
		\includegraphics[width=\textwidth]{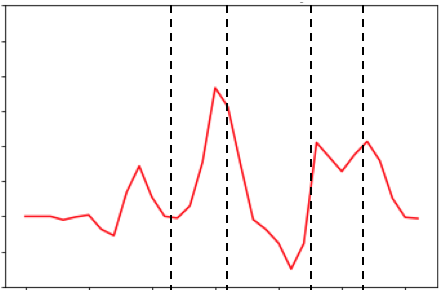}	
		{(b) section 2}\medskip
		\label{fig:hist_2}
	\end{minipage}
	\hfill
	\begin{minipage}[t]{0.32\linewidth}
		\centering
		\includegraphics[width=\textwidth]{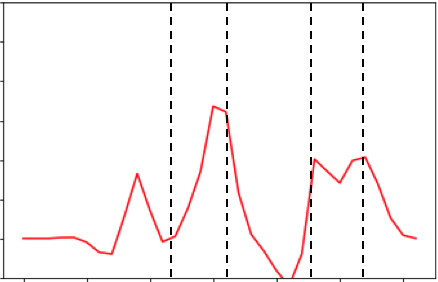}	
		{(c) section 3}\medskip
		\label{fig:hist_3}
	\end{minipage}
	\caption{A horizontal slice of generated image intensity using principle
		components from different sections, where four dotted lines in each
		subfigure correspond to the same spatial locations.} \label{fig:histograms}
\end{figure}

In order to understand the role of principal components in each section,
we generate images using principal components in each section only and
show the result in Fig.  \ref{fig:diff_sec}, where sections 1, 2 and 3
contain principal component indices 1-8, 9-120 and 121-180,
respectively.  Furthermore, for generated images in Fig.
\ref{fig:diff_sec}, we plot the intensity of a horizontal slice for the
corresponding images in Fig.  \ref{fig:histograms}, where the horizontal
slice is chosen to cross the upper circle of digit 8. We observe a
different number of peaks in these three figures. For example, there are
only two peaks in Fig. 5(a) representing two peaks in Fig. 4(a).  Figs.
5(b) and 5(c) have four peaks, representing four cross points of images
in Figs. 4(b) and 4(c), respectively. 

Based on the above analysis, we can explain the role of principal
components in each section. The first section shapes the main structure
of a digit image. The second section enhance the boundary region of the
main structure. The third and fourth sections focus on the background,
which can be discarded safely. We see that there is a trade-off in the
choice of principle components in the second section. If we want to have
simpler and clearer stroke images, it is desired to drop components in
the second section.  However, the variation of generated images will be
more limited. It is a trade-off between image quality and image
diversity. 

\subsection{Comparison with GAN and VAE}\label{sec:experi}

\begin{figure}
	\centering
	\begin{minipage}[b]{.3\linewidth}
		\centering
		\includegraphics[height=10cm]{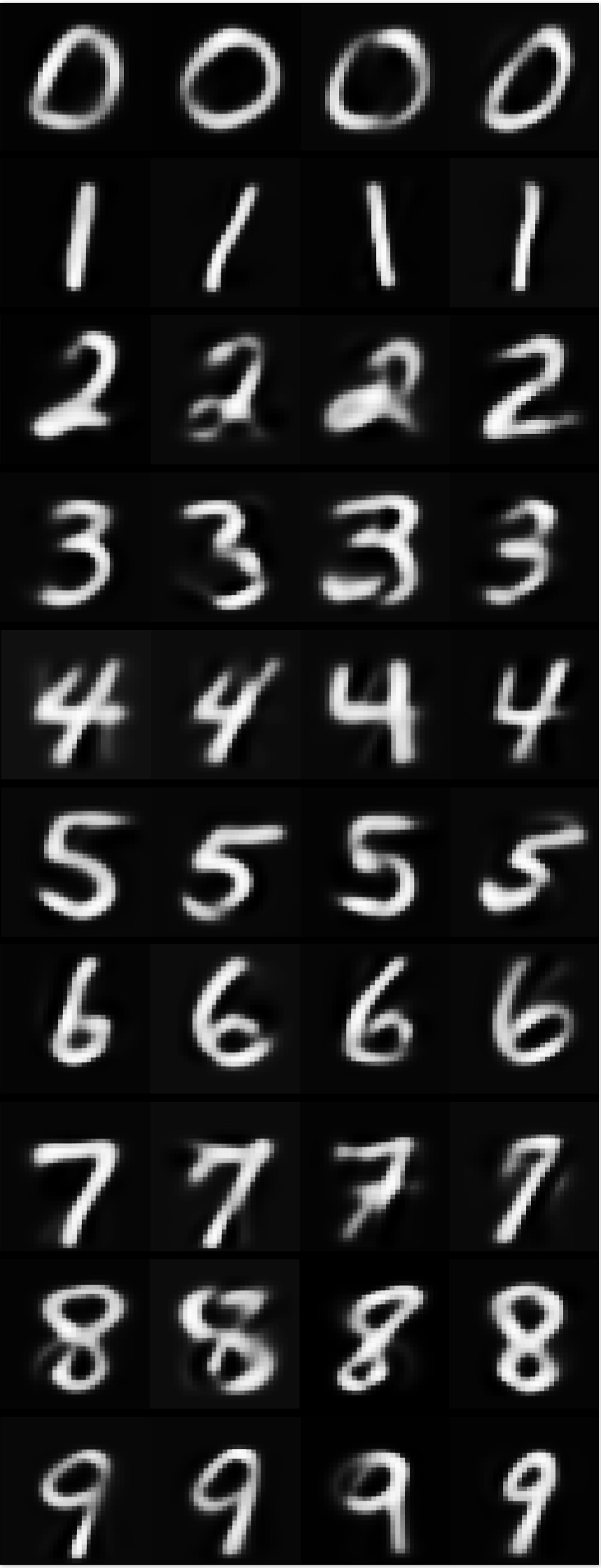}
		{(a) Our method}\medskip
		\label{fig:ourmethod}
	\end{minipage}
	\hfill
	\begin{minipage}[b]{.3\linewidth}
		\centering
		\includegraphics[height=10cm]{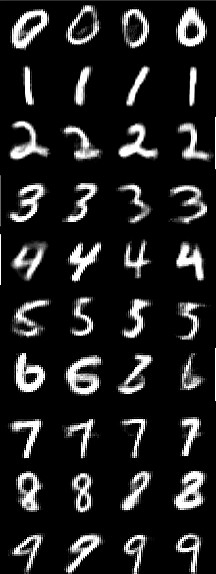}
		{(b) VAE}\medskip
		\label{fig:vae}
	\end{minipage}
	\hfill
	\begin{minipage}[b]{.3\linewidth}
		\centering
		\includegraphics[height=10cm]{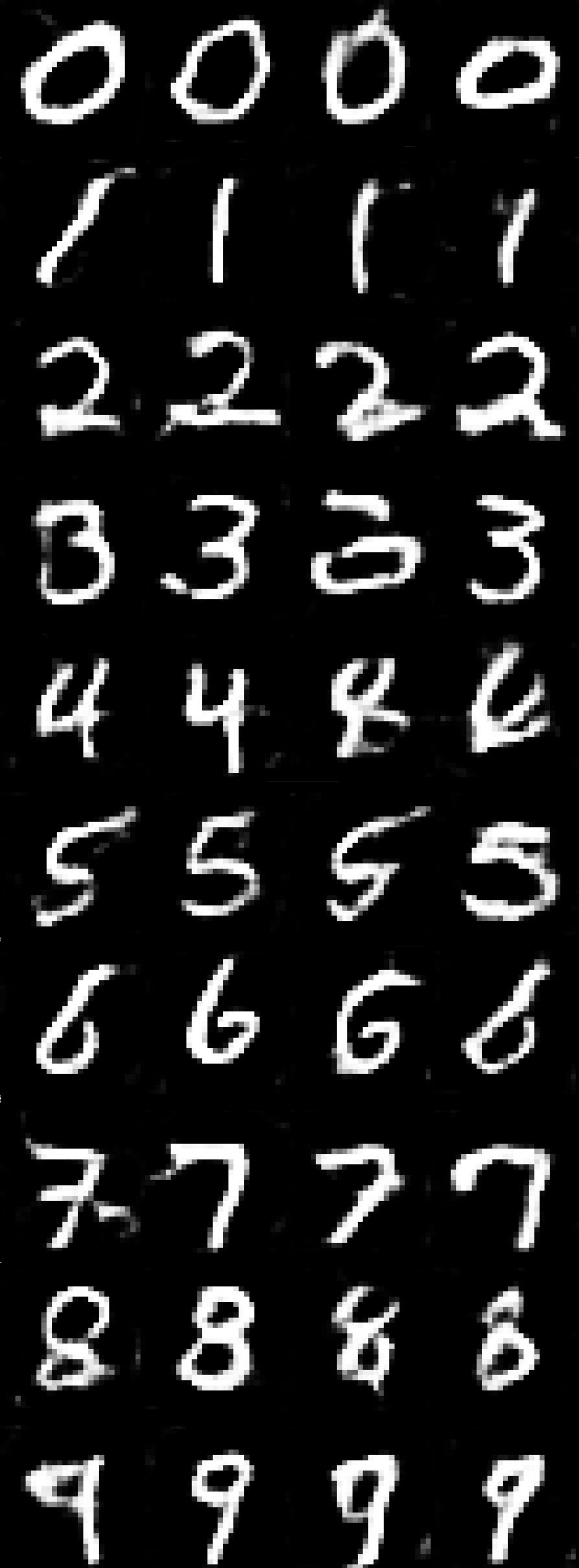}
		{(c) GAN}\medskip
		\label{fig:gan}
	\end{minipage}
	\caption{Comparison of synthesized digits using our method, the VAE and
		the GAN.}\label{fig:compare}
\end{figure}

We validate the proposed image synthesis method using the MNIST
handwritten digit dataset. The MNIST dataset is a collection of
handwritten digits from 0 to 9. We compare the synthesis results using
the proposed method, the VAE and the GAN in Fig. \ref{fig:compare}.  As
shown in the figure, we see that our method can generate images that are
different from training data with sufficient variations.  There is no
obvious difference between images synthesized by the three methods. This
indicates that our method can perform equally well as the VAE and the
GAN but at a lower training cost with a transparent design
methodology. 

\section{Discussion}\label{sec:comparative}

There are similarities between the proposed generative model and the
GAN.  First, they both use random vectors for image synthesis.  Second,
they both use convolutional operations to generate responses and feed
them to the next stage. On the other hand, there exist major differences
between them as discussed below. 

\begin{itemize}
	\item {\bf Theoretical Foundation.} It is difficult to explain GAN's
	underlying operational mechanism \cite{soltanolkotabi2018theoretical,
		wiatowski2018mathematical}. As compared with the GAN, our proposed
	method is totally transparent. 
	
	\item {\bf Model Comparison.} There are a generator and a discriminator
	in a GAN. They function as adversaries in the training process. Besides,
	the BP process is used to determine the model parameters (or filter
	weights) of both networks.  In contrast, our solution does not have the
	BP dataflow in the training.  The training process is FF and one pass. 
	
	\item {\bf Kernel Determination.} The filter weights of GANs and VAEs
	are equivalent to the transformation kernels in our proposed systme.
	These kernels are determined by the PCA of the corresponding inputs.
	Our solution is data-centric (rather than system-centric). No
	optimization framework is used in our solution. 
	
\end{itemize}

\section{Conclusion and Future Work}\label{sec:conclusion}

An interpretable generative model was proposed to synthesize handwritten
digits in this work. The multi-stage PCA system was adopted to generate
images of high resolution to overcome the drawback of the single-stage
PCA system. We discussed how to determine the principle component number
in constructing the best linear approximation subspace to the original
input space. Also, we presented two methods to detect outliers in the
synthesis procedure to improve overall image quality. It was
demonstrated by experimental results that our method can offer high
quality images that are comparable to those obtained by GAN and VAE at a
much lower training complexity since no BP is adopted. 

There are several possible directions for further exploration. First, we
may design and interpret the VAE based on the proposed framework. The
VAE has an encoder and a decoder that have structures similar to ours
They correspond the forward transform and the inverse transform
respectively. Second, it is inspiring to apply our proposed method to
face image synthesis, which is more challenging as compared to
handwritten digits. Third, it is worthwhile to study an automatic and
effective way to differentiate synthesized and real images. This is
critical to image forensic applications. 
\vfill\pagebreak

\section*{References}
\bibliographystyle{elsarticle-num} 
\bibliography{yao}

\end{document}